\documentclass[10pt,twocolumn,letterpaper]{article}

\usepackage[pagenumbers]{cvpr} % To force page numbers, e.g. for an arXiv version

\definecolor{cvprblue}{rgb}{0.21,0.49,0.74}
\usepackage[pagebackref,breaklinks,colorlinks,allcolors=cvprblue]{hyperref}
\def\paperID{9272} % *** Enter the Paper ID here
\def\confName{CVPR}
\def\confYear{2025}

\title{Break a Lag:  Triple Exponential Moving Average for Enhanced Optimization} 

\author{
    Roi Peleg\textsuperscript{1}, Yair Smadar\textsuperscript{1}, Teddy Lazebnik\textsuperscript{2,3}, Assaf Hoogi\textsuperscript{1}\\
    \textsuperscript{1}The School of Computer Science, Ariel University, Israel\\
    \textsuperscript{2}Department of Mathematics, Ariel University, Israel\\
    \textsuperscript{3}%Department of Cancer Biology, Cancer Institute, 
    University College London, London, UK\\
    {\tt\small %roi2307@gmail.com, yair.smadar1@gmail.com, tedil@ariel.ac.il, 
    assafh@ariel.ac.il}
}

\usepackage{amsmath,amsfonts}
\usepackage{algpseudocode}
\usepackage[linesnumbered,ruled]{algorithm2e}
\usepackage{epstopdf}

\usepackage{array}
\usepackage{subcaption}
\expandafter\def\csname ver@subfig.sty\endcsname{}

\usepackage{textcomp}
\usepackage{stfloats}
\usepackage{subfig}

\usepackage{url}
\usepackage{verbatim}
\usepackage{graphicx}
\usepackage{graphics}
\usepackage{float}
\usepackage{times}  % DO NOT CHANGE THIS
\usepackage{helvet}  % DO NOT CHANGE THIS
\usepackage{courier}  % DO NOT CHANGE THIS
\usepackage{natbib}  % DO NOT CHANGE THIS AND DO NOT ADD ANY OPTIONS TO IT
\usepackage{caption} % DO NOT CHANGE THIS AND DO NOT ADD ANY OPTIONS TO IT
\usepackage[utf8]{inputenc}
\usepackage{multirow}
\usepackage{amssymb}
\usepackage{epstopdf}
\usepackage{algpseudocode}
\usepackage{xcolor}
\usepackage[section]{placeins}
\usepackage{url}            % simple URL typesetting
\usepackage{booktabs}       % professional-quality tables
\usepackage{amsfonts}       % blackboard math symbols
\usepackage{nicefrac}       % compact symbols for 1/2, etc.
\usepackage{microtype}      % microtypography
\usepackage[linesnumbered,ruled]{algorithm2e}
\newcolumntype{P}[1]{>{\centering\arraybackslash}p{#1}}
\usepackage{alphabeta}
\usepackage{xcolor}

\usepackage{sidecap, caption}
\usepackage{newfloat}
\usepackage{listings}
\usepackage{lettrine}
\DeclareUnicodeCharacter{2212}{-}

\newcommand{\TEMA}{\text{TEMA}}
\newcommand{\DEMA}{\text{DEMA}}

\newcommand{\EMA}{\text{EMA}}
\newcommand{\FAME}{\text{FAME}}

\begin{document}
\maketitle

\begin{abstract}
The performance of deep learning models is critically dependent on sophisticated optimization strategies. While existing optimizers have shown promising results, many rely on first-order Exponential Moving Average (EMA) techniques, which often limit their ability to track complex gradient trends accurately. This fact can lead to a significant lag in trend identification and suboptimal optimization, particularly in highly dynamic gradient behavior.
To address this fundamental limitation, we introduce Fast Adaptive Moment Estimation (FAME), a novel optimization technique that leverages the power of Triple Exponential Moving Average. By incorporating an advanced tracking mechanism, FAME enhances responsiveness to data dynamics, mitigates trend identification lag, and optimizes learning efficiency. 
Our comprehensive evaluation encompasses different computer vision tasks including image classification, object detection, and semantic segmentation, integrating FAME into 30 distinct architectures ranging from lightweight CNNs to Vision Transformers. Through rigorous benchmarking against state-of-the-art optimizers, FAME demonstrates superior accuracy and robustness. Notably, it offers high scalability, delivering substantial improvements across diverse model complexities, architectures, tasks, and benchmarks.
\end{abstract}

\section{Introduction}
Optimization techniques play a pivotal role in advancing computer vision by enabling models to learn efficiently from data through loss function minimization. As the complexity of visual tasks—such as object detection, image segmentation, and fine-grained classification—continues to escalate, the ability of optimizers to balance computational efficiency with performance accuracy becomes increasingly paramount. Optimization methods for deep learning have evolved by addressing specific challenges in computer vision training.

Stochastic Gradient Descent (SGD) \cite{robbins1951stochastic} laid the foundation for modern optimization techniques by scaling gradients uniformly across all parameters. A significant advancement came with the introduction of Momentum \cite{Qian1999OnTM} to SGD, facilitating smoother descent trajectories and accelerating convergence by accumulating past gradient information. However, the computational intensity of SGD, particularly in the high-dimensional spaces characteristic of computer vision problems, remained a critical obstacle. To address this issue, researchers developed adaptive learning rate methods, which use nonuniform step sizes to scale the gradient during training \cite{DEMA}. These adaptive first-order methods compute individualized learning rates for each parameter, offering improved convergence in complex loss landscapes. Notable examples include AdaDelta \cite{zeiler2012adadelta}, which utilizes a moving window of past gradient updates; AdaGrad \cite{adagrad}, which normalizes gradients based on their historical magnitudes; RMSProp \cite{Tieleman2012}, which addresses AdaGrad's diminishing learning rate issue; and Adam \cite{adam}, which combines the strengths of RMSProp with momentum by applying an exponential moving average (EMA) to past gradients and squared gradients.
Subsequent variants have further refined these approaches. AMSGrad \cite{AMSGrad} ensures a non-increasing step size to enhance stability in non-convex optimization landscapes. AdaBound \cite{AdaBound} dynamically bounds learning rates to mitigate the convergence issues of adaptive methods. AdamW \cite{AdamW} decouples weight decay from the optimization step, improving generalization in image classification tasks. AdaHessian \cite{AdaHessian} incorporates second-order information to navigate complex loss surfaces more effectively.\newline
Despite faster convergence, EMA-based optimizers often track gradients non-optimally, compromising performance and generalization compared to SGD in vision tasks. This generalization gap has prompted research into hybrid approaches, such as SWATS \cite{keskar2017improving}, which alternates between SGD and Adam, and Mixing Adam and SGD (MAS) \cite{AdamSGD}, which combines their update rules using constant factors. However, these fusion strategies have shown limited performance gains, failing to fully bridge the gap between adaptive methods' fast initial progress and SGD's superior generalization.

\noindent We propose FAME, a novel optimizer leveraging Triple Exponential Moving Average (TEMA) to address the fundamental limitations of current first-order EMA-based optimization methods. \newline Our key contributions include:

\begin{itemize}
    \item \textbf{High-Order Optimization Framework:} FAME is the first optimizer to leverage Triple Exponential Moving Average (TEMA) for deep learning optimization. Unlike traditional first-order EMA-based methods, FAME's hierarchical multi-level structure enables improved optimization.
    
    \item \textbf{Addressing Limitations of traditional EMA-based Optimizers:} FAME introduces a better mechanism for handling gradient dynamics through precise gradient tracking with minimal lag, stable training in noisy environments, and accelerated convergence for large-scale vision tasks.
    
    \item \textbf{Comprehensive Validation:} Experiments across 30 neural models, a diversity of vision tasks and five public benchmarks demonstrate FAME's consistent superiority in accuracy, robustness, and training efficiency - compared with commonly used optimizers.\newline
\end{itemize}

\iffalse
\begin{itemize}
    \item \textbf{High-Order EMA-based Optimizer:} FAME introduces a TEMA-based optimizer, advancing beyond the inherent limitations of first-order EMA-based methods. Through its hierarchical multi-level exponential smoothing structure, FAME achieves simultaneous rapid gradient adaptation and robust noise reduction - a capability unattainable with the basic first-order EMAs.
    
    \item \textbf{Breaking Through First-Order Limitations:} FAME enhances optimization through precise gradient tracking with minimal lag, stable training in noisy environments, and accelerated convergence for large-scale vision tasks.
    
    \item \textbf{Comprehensive Validation:} Experiments across 30 neural models, a diversity of vision tasks and five public benchmarks demonstrate FAME's consistent superiority in accuracy, robustness, and training efficiency - compared with commonly used optimizers.\newline
\end{itemize}
\fi 

\section{FAME: Fast-Adaptive Moment Estimation}
\label{sec:fame}

\subsection{EMA: A Foundation for Trend Analysis}

EMA enables temporal data analysis. At its core, \EMA{} implements a theoretically grounded weighting scheme that decays exponentially over time, and balances noise reduction and trend preservation. This mathematical foundation sets it apart from conventional smoothing operators, particularly Gaussian kernels, by minimizing temporal lag while maintaining signal fidelity. \newline
\noindent Formally, let $x=x_{1:t} \in \mathbb{R}^t$ represents a temporal sequence up to time step $t$. EMA can be elegantly expressed through a recursive formulation:

\begin{equation}
\begin{split}
\label{EMA}
\text{EMA}(x_{1:t}) &= \beta \cdot \text{EMA}(x_{1:t-1}) + (1-\beta) \cdot x_t \\
&= (1-\beta)\sum_{i=0}^{t} \beta^i x_{t-i},
\end{split}
\end{equation}

\noindent EMA relies on its smoothing coefficient $\beta \in [0,1]$, which forces a critical trade-off: small $\beta$ values allow quick adaptation but amplify noise, while large $\beta$ values reduce noise but introduce significant delays. This means first-order EMA must always sacrifice either speed or stability.

\noindent Although first-order EMA's exponentially decaying memory structure helps stabilize gradient estimates, it fundamentally suffers from two problems: it lags behind actual changes and tends to over-smooth important gradient patterns. These issues can significantly impact optimization performance, especially in complex deep learning tasks.

\begin{figure}[h!]
\centering
  \includegraphics[width=1.1\linewidth]{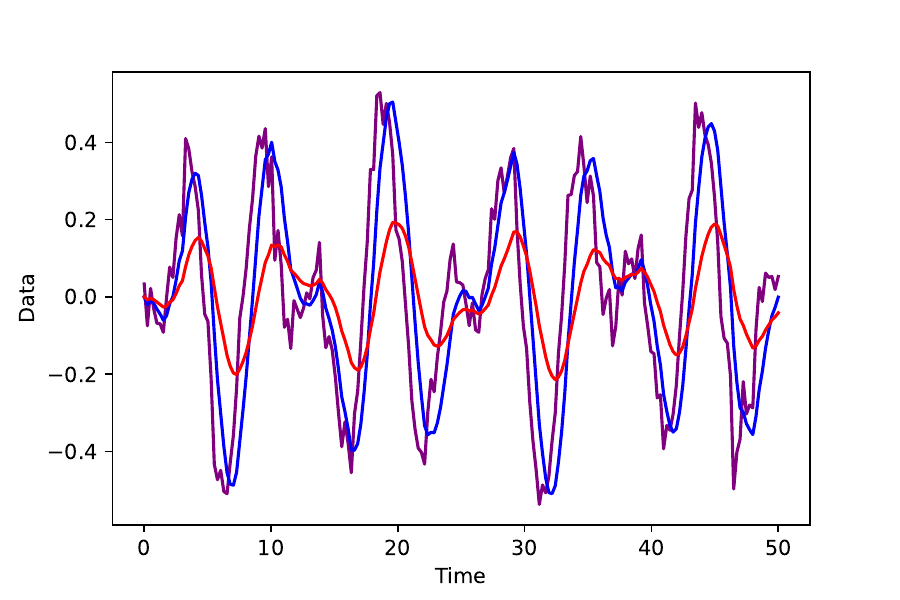}
  \caption{Simulated demonstration of gradient trend estimation and lagging. Ground truth (GT, purple), triple exponential moving average (TEMA)-based estimation (blue), and exponential moving average (EMA)-based estimation (red).}
  \label{fig:gradient_estimation}
\end{figure}

\noindent Higher-order EMA techniques systematically address these limitations. By using multiple levels of exponential smoothing, they can achieve both quick adaptation and noise reduction - capabilities that are mathematically impossible with first-order EMA's single smoothing coefficient (illustrated in Figure \ref{fig:gradient_estimation}).

\subsection{Double Exponential Moving Average (DEMA)}
We introduce advanced smoothing techniques that address the inherent lag limitations of traditional Exponential Moving Averages (EMA). The Double Exponential Moving Average (DEMA) incorporates a lag-correction mechanism through:
\begin{equation}
    \text{DEMA}(x) = 2\text{EMA}_1(x) - \text{EMA}_2(x)
    \label{eq:dema}
\end{equation}
\noindent where $\text{EMA}_k(x)$ represents k recursive applications of EMA:
\begin{equation}
    \text{EMA}_k(x) = \underbrace{\text{EMA}(\text{EMA}(\cdots \text{EMA}(x)))}_{k\text{ times}}
    \label{eq:ema_k}
\end{equation}
\noindent The key innovation in DEMA lies in its lag-correction term $[\text{EMA}_{1}(x) - \text{EMA}_2(x)]$, which provides a robust estimation of the temporal displacement between raw data $x$ and its primary smooth estimate ${EMA}_{1}(x)$. This formulation enables DEMA to achieve superior temporal responsiveness while preserving signal fidelity.

\subsection{Triple Exponential Moving Average (TEMA)}
Building upon DEMA's foundation, the Triple Exponential Moving Average (TEMA) introduces a second-order correction mechanism:
\begin{equation}
    \text{TEMA}(x) = 3\text{EMA}_1(x) - 3\text{EMA}_2(x) + \text{EMA}_3(x)
    \label{eq:tema}
\end{equation}
\noindent TEMA formulation emerges from a hierarchical lag correction process:
\begin{align}
    &\text{TEMA}(x) = \text{EMA}_1(x) 
    + [\text{EMA}_1(x) - \text{EMA}_2(x)] \nonumber \\
    &+ [(\text{EMA}_1(x) - \text{EMA}_2(x)) - \text{EMA}(\text{EMA}_1(x) - \text{EMA}_2(x))] \nonumber \\
    &= \text{EMA}_1(x) 
    + [\text{EMA}_1(x) - \text{EMA}_2(x)] \nonumber \\
    &+ [\text{EMA}_1(x) - 2\text{EMA}_2(x) 
    + \text{EMA}_3(x)] \nonumber \\
    &= 3\text{EMA}_1(x) - 3\text{EMA}_2(x) 
    + \text{EMA}_3(x)
    \label{eq:tema_derivation}
\end{align}

The formulation consists of three critical components: (1) the initial smooth estimate $\text{EMA}_1(x)$, (2) the first-order lag correction $[\text{EMA}_1(x) - \text{EMA}_2(x)]$, and (3) the second-order refinement term $[\text{EMA}_1(x) - 2\text{EMA}_2(x) + \text{EMA}_3(x)]$. This hierarchical structure enables TEMA to achieve unprecedented adaptability to temporal dynamics while maintaining signal coherence.

\begin{figure}[h!]
\centering
 \includegraphics[width=1.1\linewidth]{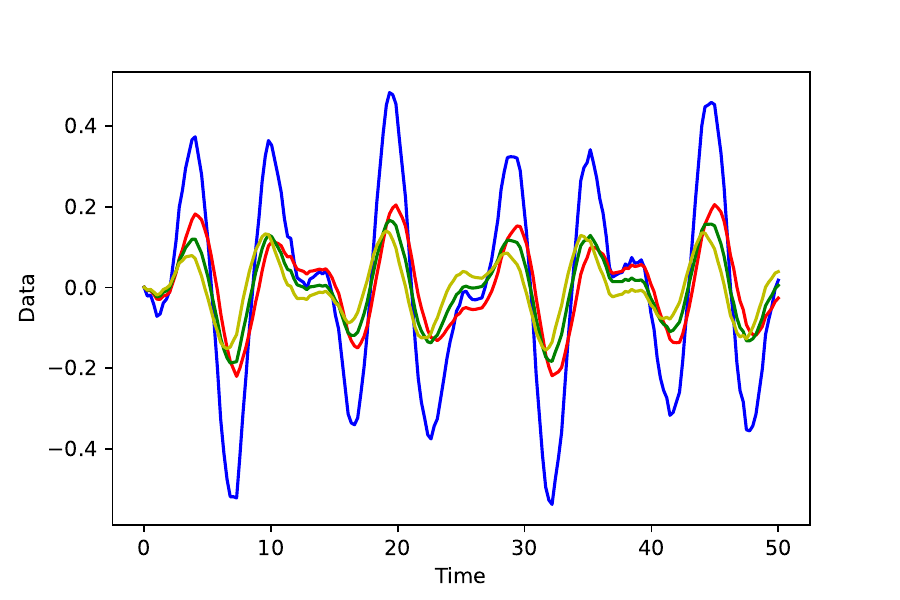}
  \caption{Decomposition of TEMA (blue) into its constituent components: $\text{EMA}_1$ (red), first-order correction $(\text{EMA}_1 - \text{EMA}_2)$ (green), and second-order correction $(\text{EMA}_1 - 2\text{EMA}_2 + \text{EMA}_3)$ (yellow). The composite TEMA signal demonstrates the synergistic integration of these components.}
\label{fig:tema_components}
\end{figure}

As visualized in Figure~\ref{fig:tema_components}, TEMA's components exhibit distinct phase characteristics that, when combined, achieve optimal lag reduction while preserving signal integrity. The complementary nature of these components enables robust temporal adaptation across diverse signal dynamics.

\subsection{Generalized k-th Order Exponential Moving Average}
We further generalize this framework to arbitrary orders through the k-th order EMA formulation:
\begin{equation}
    \text{KEMA}_k(x) = \sum_{i=1}^k (-1)^{i+1} \binom{k}{i} \text{EMA}_i(x)
    \label{eq:kema}
\end{equation}
This generalization provides fine-grained control over the temporal responsiveness-smoothness trade-off, enabling optimal adaptation to specific optimization scenarios. The binomial coefficients in this formulation ensure proper weighting of higher-order corrections, maintaining mathematical consistency with lower-order variants.\newline

\subsection{FAME Intuition}
\noindent This section provides better intuition and deeper insights into FAME's mechanisms and advantages, building upon the presented mathematical foundations.\newline
EMA is a fundamental smoothing technique that weights sequential data, assigning higher importance to recent values while exponentially diminishing the influence of older ones. This creates a local window that naturally emphasizes current patterns while maintaining historical context. When applied recursively—calculating EMA on an already smoothed signal (EMA(EMA))—the smoothing effect compounds hierarchically. The first-level EMA captures immediate trends based on its smoothing factor, while subsequent EMAs further refine this signal, effectively expanding the observation window and enabling more sophisticated pattern detection.
Building upon this foundation, FAME introduces three key innovations:
\begin{enumerate}
\item \textbf{Advanced Multi-Scale Optimization:}
Traditional first-order EMA optimizers face a fundamental speed-stability dilemma: shorter windows offer quick adaptation but amplify noise, while longer windows ensure stability at the cost of responsiveness. FAME transcends this limitation through its hierarchical triple-EMA structure. The three interconnected EMA levels form a cascading filtering pyramid, where each level operates at progressively longer time scales: the first captures immediate gradient dynamics, the second aggregates emerging patterns, and the third maintains stable long-term momentum.

   \item \textbf{Smarter Trend Recognition:} 
   First-order EMA simply follows past gradients data, making it slow to recognize new trends. TEMA's three-level mechanism acts like a verification chain: the first level spots potential changes, the second level confirms them, and the third level validates their importance. This helps it quickly identify real trends while ignoring temporary fluctuations.

   \item \textbf{Quick Error Correction:} 
   When first-order EMA makes a mistake, it takes time to recover. TEMA's structure allows for faster error correction because each level helps compensate for errors in the others.\newline
\end{enumerate}

\noindent FAME's nature makes Triple EMA especially valuable in deep learning optimization, where we need to quickly and accurately respond to changes in the model's behavior while maintaining stable training. 

\subsection{Formulation of FAME}

Let $f(\theta)$ be a differentiable stochastic scalar function that depends on the parameters $\theta$. It is a noisy version of the expected objective function $F(\theta)=\mathbb{E}[f(\theta)]$. Our goal is to minimize $F(\theta)$ with respect to its parameters $\theta$, given only a sequence $f_1(\theta), \ldots, f_T(\theta)$ of realizations of the stochastic function $f$ at subsequent time steps $1, \ldots, T$. The stochasticity in $f$ may originate from evaluations on random subsamples of data points or from inherent function noise.

\subsubsection{Gradient Estimation}

The gradient, i.e., the vector of partial derivatives of $f_t$ with respect to $\theta$, evaluated at time step $t$, is denoted by:
\begin{equation}
\label{eq:gradient}
g_t = \nabla_{\theta} f_t(\theta)
\end{equation}
\subsubsection{Moment Estimation}

Similar to Adam, FAME tracks the first-order and second-order moments $(m_t, v_t)$ of the gradients using an exponential moving average (EMA):
\begin{equation}
\label{eq:moments}
\begin{aligned}
m_t &= \beta_1 m_{t-1} + (1 - \beta_1) g_t \\
v_t &= \beta_2 v_{t-1} + (1 - \beta_2) g_t^2
\end{aligned}
\end{equation}
where the hyperparameters $\beta_1, \beta_2 \in [0, 1)$ control the exponential decay rates of these moving averages.

\noindent FAME extends beyond traditional optimizers by incorporating higher-order moment estimations. It tracks additional variables $(dm_t, dv_t)$ for estimating $\text{EMA}_2$:
\begin{equation}
\label{eq:second_order_moments}
\begin{aligned}
dm_t &= \beta_3 dm_{t-1} + (1 - \beta_3) m_t \\
dv_t &= \beta_4 dv_{t-1} + (1 - \beta_4) v_t
\end{aligned}
\end{equation}
Based on $(dm_t, dv_t)$, FAME further estimates $\text{EMA}_3$ using variables $(tm_t, tv_t)$:
\begin{equation}
\label{eq:third_order_moments}
\begin{aligned}
tm_t &= \beta_5 tm_{t-1} + (1 - \beta_5) dm_t \\
tv_t &= \beta_4 tv_{t-1} + (1 - \beta_4) dv_t
\end{aligned}
\end{equation}
Here, $(m_0, v_0)$, $(dm_0, dv_0)$, and $(tm_0, tv_0)$ are all initialized to 0.\newline

\noindent Incorporating Eqs. (\ref{eq:moments}-\ref{eq:third_order_moments}) into the Triple Exponential Moving Average (TEMA) equation (Eq. \ref{eq:tema}) yields:

\begin{equation}
\label{eq:fame_moments}
\begin{aligned}
m_{\text{FAME}_t} &= 3m_t - 3dm_t + tm_t \\
v_{\text{FAME}_t} &= 3v_t - 3dv_t + tv_t
\end{aligned}
\end{equation}

\subsubsection{Parameter Update Rule}

The final parameter update equation for FAME is given by:

\begin{equation}
\label{eq:fame_update}
\theta_t = \theta_{t-1} - \alpha \cdot \frac{m_{\text{FAME}_t}}{\sqrt{v_{\text{FAME}_t} + \epsilon}}
\end{equation}

where $\alpha$ is the learning rate and $\epsilon$ is a small constant to prevent division by zero.

\section{Experiments}
We validated the FAME optimizer across five diverse benchmarks and 30 model models with varying complexities, covering tasks like detection, classification, and semantic understanding. FAME was compared against six popular optimizers: SGD + Momentum, Adam, Adagrad, AdamW, AdaBound, and AdaHessian.

\begin{algorithm}[ht!]
\caption{FAME Optimizer}
\label{alg:fame-optimizer}

\KwIn{Hyper-parameters: $\beta_1, \beta_2, \beta_3, \beta_4, \beta_5, \alpha, \epsilon$}
\KwOut{Optimized parameters $\hat{\theta}_t$ after convergence}

% Initialization
\textbf{Initialize:} $m_0 \gets 0$, $v_0 \gets 0$, $dm_0 \gets 0$, $dv_0 \gets 0$, $tm_0 \gets 0$, $tv_0 \gets 0$, $t \gets 0$

\While{$\theta_t$ not converged}{
    $t \gets t + 1$
    
    % First moment estimate
    $m_t \gets \beta_1 m_{t-1} + (1-\beta_1) g_t$
    
    % Second moment estimate
    $v_t \gets \beta_2 v_{t-1} + (1-\beta_2) g_t^2$
    
    % First moment difference
    $dm_t \gets \beta_3 dm_{t-1} + (1-\beta_3) m_t$
    
    % Second moment difference
    $dv_t \gets \beta_4 dv_{t-1} + (1-\beta_4) v_t$
    
    % Triple moment for first estimate
    $tm_t \gets \beta_5 tm_{t-1} + (1-\beta_5) dm_t$
    
    % Triple moment for second estimate
    $tv_t \gets \beta_4 tv_{t-1} + (1-\beta_4) dv_t$
    
    % FAME estimates
    $m_{\text{FAME}_t} \gets 3m_t - 3dm_t + tm_t$
    $v_{\text{FAME}_t} \gets 3v_t - 3dv_t + tv_t$
    
    % Update rule
    $\theta_t \gets \theta_{t-1} - \alpha \cdot \frac{m_{\text{FAME}_t}}{\sqrt{v_{\text{FAME}_t}} + \epsilon}$
}

\Return{$\theta_t$}
\end{algorithm}

\subsection{\textbf{Experimental Data}}

We tested our proposed FAME on five public benchmarks:
CIFAR-100 \cite{krizhevsky2009learning}, MS-COCO \cite{lin2014microsoft}, PASCAL-VOC \cite{Everingham15}, Cityscapes \cite{cityscapes}, and the large scale ImageNet \cite{ILSVRC15}.

\subsection{\textbf{Implementation Details}}
\noindent To ensure a rigorous comparison, all optimizers were evaluated under identical initialization conditions and trained from scratch. While this approach may yield lower absolute accuracies compared to pre-trained models, it enables an unbiased assessment of each optimizer's intrinsic capabilities. For fair comparison, we conducted comprehensive hyperparameter tuning for each optimizer, starting from literature-recommended defaults and employing grid search to identify optimal settings. Given FAME's extensive validation across 30 models, we follow standard practice by referencing the established hyperparameter configurations from prior work for each model.

\iffalse
\begin{table}[H]
\centering
\begin{tabular}{|P{1.5cm}|P{0.85cm}|P{0.85cm}|P{1cm}|P{0.85cm}|P{1.1cm}|}\hline
  &CIFAR & COCO & PASCAL & City & ImageNet\\\hline\hline
   Batch size & 128&128&48&16&128\\\hline
   LR & 0.001&0.001&0.0334&0.001&0.001\\\hline
   Momentum & 0.94 &0.94 &0.75 &0.9 &0.9\\\hline
   WD & 0.005 &0.005 &0.00025 &0.0001 &0.0001 \\\hline
   Warmup & 3&3&3&3&3\\\hline
   Epochs & 200&250&200&170&40\\\hline
\end{tabular}

\caption{VALUES OF HYPER-PARAMETERS. LR = LEARNING RATE; WD = WEIGHT DECAY, WARM-UP = WARM-UP EPOCHS}
\label{Hyper}
\end{table}
\fi

\section{Results}
In pursuit of a comprehensive analysis, we computed several statistical metrics, including accuracy, mAP (mean Average Precision), precision, recall, and F1-score. This involved evaluating the average and standard deviation across three distinct model initializations, solidifying the robustness of our method's validation.

\subsection{\textbf{Image Classification}}
\subsubsection{CIFAR-100}
FAME demonstrates superior performance on CIFAR-100 across 17 diverse models (Table \ref{table1}). When compared with commonly used optimizers, FAME achieved superior accuracy in 84.6\% of the models, consistently outperforming existing methods with averaged accuracy improvements of 1.16\% over AdamW, 1.57\% over AdaBound, 3.52\% over AdaHessian, and 16.33\% over AdaGrad.

\noindent Beyond accuracy gains, FAME demonstrated superior training stability across diverse models. For CNN-based models, FAME reduced epoch-wise accuracy variance by 29.27\% compared with existing optimizers (Adam, SGD, AdamW, AdaBound, AdaHessian, AdaGrad). This stability advantage persisted in transformer-based models, where FAME achieves a 21.54\% variance reduction compared with the same baseline optimizers. Such consistent stability across different models highlights FAME's robustness to gradient dynamics, effectively mitigating the training fluctuations common in current methods.

\iffalse
\begin{figure*}[ht!]
  \centering
  \begin{subfigure}[b]{0.39\linewidth}
    \centering
    \includegraphics[width=\linewidth]{effb3comp.pdf}
    \caption{}
    \label{OptimizersCompare}
  \end{subfigure}%
  \hfill
  \begin{subfigure}[b]{0.305\linewidth}
    \centering
    \includegraphics[width=\linewidth]{CIFARCompare.png}
    \caption{}
    \label{CIFARCompare}
  \end{subfigure}%
  \hfill
  \begin{subfigure}[b]{0.295\linewidth}
    \centering
    \includegraphics[width=\linewidth]{ImageNetCompare.png}
    \caption{}
    \label{ImageNet}
  \end{subfigure}
  \caption{Performance comparison of FAME and other optimizers across various models and datasets: (a) EfficientNet-B3 on CIFAR-100, (b) RevVit on CIFAR100, and (c) ResNet-18 on ImageNet.}
  \label{fig:optimizer_comparison}
\end{figure*}
\fi

\iffalse
\begin{figure}[ht!]
  \includegraphics[width=1.1\linewidth]{effb3comp.pdf}
  \centering
\caption{FAME vs.\ other optimizers on CIFAR-100. EfficientNet-B3 model is trained from scratch.}
\label{OptimizersCompare}
\end{figure} 

\begin{figure}[ht!]
 \centering
\includegraphics[width=0.7\linewidth]{CIFARCompare.png}
  \caption{Comparison of the performance stability of FAME, Adam, and SGD on CIFAR100 using RevVit transformer model.}
    \label{CIFARCompare}
\end{figure}
\fi
\subsubsection{ImageNet Benchmark}
Our ImageNet experiments reveal FAME's compelling advantages across models. For CNN-based models, FAME achieved consistent accuracy improvements of 0.8\% over Adam and 0.65\% over SGD. The gains were even more pronounced in Transformer-based models, where FAME surpassed Adam by 0.97\% and SGD by 1.27\%.
Beyond accuracy improvements, FAME enhanced the training process in two critical aspects. First, it provided superior training stability, reducing accuracy fluctuations variance by 5.64\% compared with SGD, Adam, and AdamW. Second, it accelerated convergence significantly, reaching 90\% of peak accuracy while using only 78\% of the epochs required by other optimizers.
This comprehensive enhancement in accuracy, stability, and training efficiency establishes FAME as a powerful optimization solution for large-scale vision tasks.

\iffalse
\begin{figure}[h!]
 \centering
\includegraphics[width=0.7\linewidth]{ImageNetCompare.png}
  \caption{Classification accuracy on ImageNet (training ResNet-18 from \textit{scratch}).}
    \label{ImageNet}
\end{figure}
\fi

\begin{table*}[h]
\caption{Image Classification. CIFAR-100 and Imagenet benchmarks. Comparison of classification accuracy (Mean $\pm$ Std) supplied by different optimizers across models. \textbf{Best results for each model are bolded}. All SGD and Adam results are comparable to the literature results when training from \underline{scratch}. Std was calculated for three different initializations. "$L_{drloc}$" signifies the addition of dense relative localization loss \cite{drloc}, which improves the robustness of visual transformers}
\centering
\begin{tabular}{|P{2cm}|P{1.9cm}|P{3.6cm}|P{2.4cm}|P{2.4cm}|P{2.4cm}|}
\hline
Dataset & Type & model & Our FAME & Adam & SGD + Momentum \\
\hline\hline 
\multirow{16}{*}{CIFAR-100} 
& \multirow{11}{*}{CNN} & EfficientNet-b3 \cite{EfficientNet}& \textbf{0.761 $\pm$ 0.008} & 0.749 $\pm$ 0.015 & 0.748 $\pm$ 0.017 \\
&  & EfficientNet-b5 \cite{EfficientNet} & \textbf{0.793 $\pm$ 0.010} & 0.776 $\pm$ 0.016 & 0.771 $\pm$ 0.017 \\
&  & MobileNetV3-Large \cite{MobileNetV3} & \textbf{0.756 $\pm$ 0.005} & 0.706 $\pm$ 0.007 & 0.744 $\pm$ 0.011 \\
&  & DenseNet-121 \cite{DenseNet} & \textbf{0.738 $\pm$ 0.008} & 0.714 $\pm$ 0.015 & 0.727 $\pm$ 0.011 \\
&  & DenseNet-201 \cite{DenseNet} & 0.739 $\pm$ 0.005 & \textbf{0.742 $\pm$ 0.006} & 0.733 $\pm$ 0.006 \\
&  & SEResNet-18 \cite{SEResNet}& \textbf{0.715 $\pm$ 0.009} & 0.704 $\pm$ 0.009 & 0.692 $\pm$ 0.016 \\
&  & SE-ResNet-50 \cite{SEResNet} & \textbf{0.762 $\pm$ 0.007} & 0.751 $\pm$ 0.009 & 0.747 $\pm$ 0.011 \\
&  & Dspike (ResNet-18) \cite{Dspike} & \textbf{0.723 $\pm$ 0.006} & 0.708 $\pm$ 0.008 & 0.693 $\pm$ 0.019 \\
&  & PyramidNet-272 \cite{PyramidNet-272}& \textbf{0.832 $\pm$ 0.006} & 0.811 $\pm$ 0.008 & 0.808 $\pm$ 0.008 \\
&  & Inception-v3 \cite{Inception-v3}& \textbf{0.751 $\pm$ 0.006} & 0.743 $\pm$ 0.008 & 0.702 $\pm$ 0.015 \\
&  & WideResNet 40-4 \cite{WideResNet}& \textbf{0.719 $\pm$ 0.007} & 0.706 $\pm$ 0.009 & 0.708 $\pm$ 0.007 \\
\cline{2-6}
& \multirow{4}{*}{Transformer} & ViT-Base+$L_{drloc}$ \cite{ViT-Base}& 0.743 $\pm$ 0.003 & 0.739 $\pm$ 0.002 & \textbf{0.775 $\pm$ 0.006} \\
&  & Swin-T+$L_{drloc}$ \cite{swin}& \textbf{0.774 $\pm$ 0.002} & 0.769 $\pm$ 0.007 & 0.762 $\pm$ 0.002 \\
&  & CvT-13+$L_{drloc}$ \cite{cvt}& \textbf{0.795 $\pm$ 0.001} & 0.794 $\pm$ 0.001 & 0.780 $\pm$ 0.005 \\
&  & T2T-ViT-14+$L_{drloc}$ \cite{T2T}& 0.781 $\pm$ 0.003 & 0.783 $\pm$ 0.001 & \textbf{0.806 $\pm$ 0.002} \\
\hline\hline
\multirow{4}{*}{ImageNet} 
& & MobileNetV3-Large \cite{MobileNetV3}& \textbf{0.729 $\pm$ 0.001} & 0.726 $\pm$ 0.001 & 0.727 $\pm$ 0.001 \\
& CNN & EfficientNet-b5 \cite{EfficientNet}& \textbf{0.785 $\pm$ 0.016} & 0.772 $\pm$ 0.014 & 0.774 $\pm$ 0.016 \\\cline{2-6}
&  & CvT-13 \cite{cvt}& \textbf{0.824 $\pm$ 0.009} & 0.820 $\pm$ 0.010 & 0.819 $\pm$ 0.016 \\
& Transformer & CAFormer-S18 \cite{CAFormer}& \textbf{0.848 $\pm$ 0.010} & 0.835 $\pm$ 0.013 & 0.833 $\pm$ 0.012 \\
& & CAFormer-S36 \cite{CAFormer}& \textbf{0.869 $\pm$ 0.010} & 0.857 $\pm$ 0.013 & 0.851 $\pm$ 0.012 \\
\hline
\end{tabular}
\label{table1}
\end{table*}

\subsection{\textbf{Object Detection}}
\subsubsection{MS-COCO Benchmark}

Table \ref{table2} demonstrates FAME's effectiveness on the MS-COCO benchmark using YOLOv5-S model, where it consistently outperforms standard optimizers including SGD, Adam, and AdamW. FAME achieved a mean Average Precision (mAP@0.5) of 0.569, surpassing SGD (0.549), Adam (0.265), and AdamW (0.446), and also excelled in mAP@0.5:0.95 with a score of 0.375 compared to 0.352 for SGD, 0.211 for Adam, and 0.271 for AdamW. In terms of Precision, FAME scored 0.663, indicating a high proportion of true positives, while Recall was at 0.521, showcasing its effectiveness in capturing relevant objects. Additionally, FAME's F1-Score of 0.583 reflects a strong balance between precision and recall, significantly outperforming SGD (0.518), Adam (0.275), and AdamW (0.471). Overall, these results affirm FAME’s enhanced object detection capabilities, positioning it as a robust optimization method in comparison to established techniques.

\begin{table}[h!]
\caption{Object Detection. MS-COCO by YOLOv5-s \cite{Yolov5-s}. Accuracy by Mean $\pm$ Std. The Std was calculated over three initializations. The results match those of training from \textbf{scratch}, as reported in the literature.}
\centering
\begin{tabular}{|P{2cm}|P{1.1cm}|P{1.1cm}|P{1.1cm}|P{1.1cm}|}\hline
 & Our FAME &    SGD &    Adam &   AdamW \\\hline\hline
  mAP@0.5 &    \textbf{0.569 $\pm$ 0.003} & 0.549 $\pm$ 0.005 & 0.265 $\pm$ 0.021 & 0.446 $\pm$ 0.013 \\\hline
  mAP@0.5:0.95 &    \textbf{0.375 $\pm$ 0.003} & 0.352 $\pm$ 0.005 & 0.211 $\pm$ 0.021 & 0.271 $\pm$ 0.014 \\\hline
  Precision &  \textbf{0.663 $\pm$ 0.009} & 0.658 $\pm$ 0.011 & 0.332 $\pm$ 0.013 & 0.542 $\pm$ 0.011  \\\hline
  Recall &    \textbf{0.521 $\pm$ 0.006} & 0.427 $\pm$ 0.006 & 0.234 $\pm$ 0.019 & 0.417 $\pm$ 0.007 \\\hline
  F1-Score &    \textbf{0.583 $\pm$ 0.008} & 0.518 $\pm$ 0.012 & 0.275 $\pm$ 0.014 & 0.471 $\pm$ 0.004 \\\hline
\end{tabular}
\label{table2}
\end{table}

\subsubsection{Pascal-VOC Benchmark}
Table \ref{table3} presents the analysis of the Pascal Visual Object Classes (VOC) dataset for detection and classification tasks, highlighting FAME's advantages over conventional optimizers like SGD, Adam, and AdamW. For the YOLOv5-S model, FAME achieved a mean Average Precision (mAP) of 0.812, which exceeds the performance of SGD (0.787), Adam (0.651), and AdamW (0.801). Similarly, FAME recorded an impressive mAP of 0.851 for the YOLOv5-m model, compared with SGD (0.828), Adam (0.662), and AdamW (0.83). In classification using the RevViT model, FAME reached an AUC score of 0.696, surpassing SGD (0.679), Adam (0.637), and AdamW (0.671). These findings demonstrate that FAME consistently enhances performance in both detection and classification tasks, underscoring its efficacy as an optimization technique relative to commonly used methods.

\begin{table}[h!]
\caption{Pascal visual object classes (VOC) detection/classification. Std was calculated over three initializations. The results match those for training from scratch as reported in literature.}
\centering
\begin{tabular}{|P{2cm}|P{1.1cm}|P{1.1cm}|P{1.1cm}|P{1.1cm}|}\hline
&Our Fame & SGD & Adam & AdamW \\\hline\hline
Yolov5-s \cite{Yolov5-s} (mAP) - \textit{Detection} & \textbf{0.812 $\pm$ 0.007} & 0.787 $\pm$ 0.007 & 0.651 $\pm$ 0.009 & 0.801 $\pm$ 0.009\\\hline
Yolov5-m \cite{Yolov5-m} (mAP) -\textit{Detection} & \textbf{0.851 $\pm$ 0.006}& 0.828 $\pm$ 0.007 & 0.662 $\pm$ 0.011 & 0.830 $\pm$ 0.009\\\hline
RevViT \cite{revit} (AUC) - \textit{Classification} & \textbf{0.696 $\pm$ 0.006} & 0.679 $\pm$ 0.007 & 0.637 $\pm$ 0.009 & 0.671 $\pm$ 0.007\\\hline
\end{tabular}
\label{table3}
\end{table}

\subsection{\textbf{Semantic Segmentation}}
\subsubsection{Cityscapes Benchmark}
Our evaluation of the Cityscapes benchmark demonstrates FAME's consistent superiority across diverse models (Table \ref{table4}). FAME achieved the highest accuracy in 6 out of 7 models (85.7\%), with improvements ranging from 0.3\% to 1\% over Adam and 2.5\% to 6.3\% over SGD. Notably, FAME showed particularly strong performance on resource-efficient models like MobileNetV2 (+0.7\% over Adam, +6.3\% over SGD) and more complex models like DL-V3+ResNet101 (+1.1\% over Adam, +4.7\% over SGD). The results also reveal FAME's stability advantage, evidenced by consistently lower standard deviations across three initializations.

\begin{table}[hbt!]
\caption{Semantic Segmentation. Cityscapes benchmark. "DL" - DeepLab \cite{DeepLab}. Comparison of classification accuracy (Mean $\pm$ Std) supplied by different optimizers across models. \textbf{Best results for each model are bolded}. All SGD and Adam results are comparable to the literature results when training from \textbf{\underline{scratch}}. Std was calculated over three different initializations.}
\centering
\begin{tabular}{|P{2.6cm}|P{1.4cm}|P{1.4cm}|P{1.4cm}|}
\hline
Model & Our FAME & Adam & SGD \\
\hline\hline
DL-V3+HRNetV2-32 & \textbf{0.633 $\pm$ 0.004} & 0.623 $\pm$ 0.007 & 0.582 $\pm$ 0.004 \\\hline
DL-V3+HRNetV2-48 & \textbf{0.641 $\pm$ 0.009} & 0.634 $\pm$ 0.011 & 0.609 $\pm$ 0.007 \\\hline
DL-V3+ResNet50* & 0.757 $\pm$ 0.009 & \textbf{0.759 $\pm$ 0.011} & 0.705 $\pm$ 0.013 \\\hline
DL-V3+ResNet101* & \textbf{0.778 $\pm$ 0.009} & 0.767 $\pm$ 0.014 & 0.731 $\pm$ 0.012 \\\hline
DL-V3+MobileNetV2* & \textbf{0.730 $\pm$ 0.013} & 0.723 $\pm$ 0.014 & 0.667 $\pm$ 0.018 \\\hline
DL-V3+Xception & \textbf{0.605 $\pm$ 0.002} & 0.602 $\pm$ 0.001 & 0.572 $\pm$ 0.008 \\\hline
HANet-ResNet101* & \textbf{0.836 $\pm$ 0.001} & 0.833 $\pm$ 0.001 & 0.817 $\pm$ 0.013 \\
\hline
\end{tabular}
\label{table4}
\end{table}

\subsection{\textbf{Robustness across Datasets, models, and Weight Initializations}}
Tables \ref{table1}-\ref{table4} show that for 87.16\% of the models, FAME outperforms all the other optimizers examined in this study, and its performance is comparable with those of the other optimizers for the remaining 12.9\% cases. Thus, its high robustness within and across datasets was established. Furthermore, FAME demonstrated a lower standard deviation across the different weight initializations.

\subsection{\textbf{Ablation Study}}

\begin{itemize}
    \item \textbf{Effect of High-Order EMAs} --- Table \ref{table5} shows that increasing the order of the EMA from simple \EMA{} to \DEMA{}, and then to \TEMA{}, leads to improved model performance. Although the $4^{th}$ order was also tested, it underperformed compared to \TEMA{}, which balanced between trend estimation, smoothness and lag reduction. The results emphasize that our triple-based FAME consistently outperformed other high-order EMAs across various datasets, tasks, and models, delivering superior accuracy.

    \item \textbf{Effect of TEMA on Different Moments} --- Table \ref{PartialFAME} illustrates how incorporating TEMA into the $1^{st}$ and $2^{nd}$ moments of FAME enhanced its performance. The shift from EMA to Partial TEMA (for $1^{st}$ $m_{\FAME{}_t}$ moment only) and then to full FAME (for both $m_{\FAME{}_t}$ and $v_{\FAME{}_t}$ moments) consistently boosted accuracy. This demonstrates FAME’s capability to improve optimization across object detection, classification, and segmentation tasks, regardless of the dataset or model.
\end{itemize}

\begin{table}[ht!]
\caption{Effect of the optimizer order}
\centering
\begin{tabular}{|P{1.4cm}|P{2cm}|P{0.9cm}|P{0.9cm}|P{0.9cm}|}
\hline
 Dataset & Model & EMA & DEMA & Our FAME \\\hline\hline
 CIFAR-100 & ResNet34 & 0.721 $\pm$ 0.008 & 0.727 $\pm$ 0.006 & \textbf{0.736 $\pm$ 0.006}\\\hline
 MS-COCO & YOLOv5-s \cite{Yolov5-s} & 0.265 $\pm$ 0.021 & 0.317 $\pm$ 0.004 & \textbf{0.569 $\pm$ 0.013}\\\hline
 PASCAL-VOC & YOLOv5-m \cite{Yolov5-m} & 0.662 $\pm$ 0.011 & 0.806 $\pm$ 0.005 & \textbf{0.851 $\pm$ 0.006} \\\hline
 Cityscapes & DL-V3+ResNet50*& 0.743 $\pm$ 0.008 & 0.751 $\pm$ 0.005 & \textbf{0.757 $\pm$ 0.005}\\\hline
\end{tabular}
\label{table5}
\end{table}

\begin{table}[ht!]
\caption{Ablation study. "Partial FAME" means using TEMA for $M_{FAME_T}$ only, while "Our FAME" means using TEMA for both $M_{FAME_T}$ and $V_{FAME_T}$. CIFAR100: ResNet50, MS-COCO: YOLOv5-s, Pascal-VOC: YOLOv5-m, and Cityscapes: ResNet50+ DeepLabV3.}
\centering
\begin{tabular}{|P{1.7cm}|P{1.6cm}|P{1.6cm}|P{1.6cm}|}
\hline 
Dataset  & Adam (Original EMA) & Partial FAME & Our FAME \\\hline\hline
{CIFAR-100 (\textit{\textbf{Accuracy}})} & 0.721 $\pm$ 0.008 & 0.729 $\pm$ 0.005 & \textbf{0.736 $\pm$ 0.006} \\\hline 
{MS-COCO (\textit{\textbf{mAP}})}& 0.265 $\pm$ 0.021 & 0.504 $\pm$ 0.006 & \textbf{0.569 $\pm$ 0.013} \\\hline
{PASCAL-VOC (\textit{\textbf{mAP}})}&  0.662 $\pm$ 0.011 & 0.829 $\pm$ 0.008 & \textbf{0.851 $\pm$ 0.006}
\\\hline
{CityScapes (\textit{\textbf{mean IoU}})} & 0.743 $\pm$ 0.008  & 0.751 $\pm$ 0.005 & \textbf{0.757 $\pm$ 0.005}\\\hline
\end{tabular}
\label{PartialFAME}
\end{table}

\subsection{\textbf{Sensitivity to Hyper-parameters Tuning}}
\subsubsection{\textbf{FAME's Hyper-parameter Robustness}}
While FAME introduces three additional parameters ($\beta_3$, $\beta_4$, $\beta_5$) beyond Adam's $\beta_1$ and $\beta_2$, it demonstrates remarkable stability. Across five major benchmarks, FAME shows only 3.72\% performance variation compared to Adam's 10.12\% sensitivity - highlighting its robustness while delivering superior optimization. Notably, these parameters remained fixed across all experiments, eliminating the need for per-task tuning.

\subsubsection{General Deep Model's Hyper-parameters}
To ensure fair evaluation, we performed comprehensive grid searches over all core hyperparameters (batch size, learning rate, momentum, weight decay, epochs) for each model, starting from established defaults and methodically exploring neighboring values - creating a robust foundation for meaningful comparisons.

\subsection{\textbf{Memory Cost and Computational Efficiency}}
Our assessment of FAME's computational demands reveals a compelling efficiency trade-off:

\begin{enumerate}
    \item \textit{Per-Epoch Performance} -- While FAME requires modestly higher resources per epoch (+7\% memory, +5\% computation time) compared to ADAM and SGD optimizers,    
    \item \textit{Overall Efficiency} -- FAME's superior convergence characteristics dramatically offset these minor costs - achieving full convergence in just 76.3\% of the training epochs required by traditional optimizers, averaged for all models and tested benchmarks.
\end{enumerate}

This effectively means that FAME makes far better use of computational resources, requiring only about one-fifth of the total training time despite slightly higher per-epoch overhead.

\subsection{\textbf{Convergence Analysis}}
To demonstrate the convergence of the proposed FAME algorithm, we begin by establishing a mathematical framework. Let \(d \in \mathbb{N}\) represent the number of parameters in the function \(F: \mathbb{R}^d \rightarrow \mathbb{R}\) that we aim to optimize. In the domain of machine learning, \(F\) embodies the complete training objective function, and optimization algorithms are employed to locate critical points within \(F\). Specifically, our focus lies on optimization techniques that expand upon the classical Gradient Descent (GD) algorithm by integrating a heavy-ball style momentum parameter \cite{teddy_3}. These algorithms exhibit a continuously diminishing step size from an initial finite step, leading to a monotonically decreasing process \cite{adam}. This behavior underscores their progression towards convergence.

Let us assume several constraints: \(0 \leq \beta_5 \leq \beta_3 \leq \beta_1 \leq 1\) and \(0 \leq \beta_4 \leq \beta_2 \leq 1\), along with the condition \(\beta_1 \leq \beta_2\), and a non-negative value for \(\alpha\).

According to Algorithm 1, we define three sequences, \(m_i, v_i, \theta_i \in \mathbb{R}^n\), where \(n\) denotes the dimension of the parameter space. Given an initial point \(\theta_0 \in \mathbb{R}^n\) and setting \(m_0 = v_0 = 0\), we proceed under three primary assumptions:

\begin{itemize}
\item{The loss function \(F\) is bounded below by an arbitrary function \(F^*\) for any point \(\forall \theta \in \mathbb{R}^n: F^*(\theta) \leq F(\theta)\).}
\item{The loss function, concerning the max-norm (\(l_\infty\)), is uniformly and surely bounded: \(\forall \epsilon > 0, \exists r > \epsilon \forall \theta \in \mathbb{R}^n: ||\nabla F||_\infty \leq r - \epsilon\).}
\item{The loss function maintains L-Lipschitz continuity concerning the \(l_2\)-norm: \(\forall \theta, \zeta \in \mathbb{R}^n: ||\nabla F(\theta) - F(\zeta)||_2 \leq L||\theta - \zeta||_2\).}
\end{itemize}

For a specific number of iterations \(N \in \mathbb{N}\), we introduce \(\tau\) as a random index ranging from \(\{0, \dots, N−1\}\), where \(\forall i \in \mathbb{N}: i < N, P[\tau = i] \propto 1 - \beta_1^{N-i}\). This implies that for \(\beta_1 = 0\), \(\tau\) is uniformly sampled, while higher orders approaching zero lead to fewer samples in later iterations. This stratagem aims to limit the expected squared gradient norm at iteration \(\tau\).

For \(N > \beta_1/(1 - \beta_1) \in \mathbb{N}\), the following inequality holds:

\begin{equation}
  \begin{aligned}
    &||\nabla F(\theta_\tau)||^2 \leq 2r\sqrt{N} \frac{F(\theta_0) - F^*}{\alpha (N- \frac{\beta_1}{1 - \beta_1})} \quad + \\
    & \frac{\sqrt{N}}{N - \frac{\beta_1}{1 - \beta_1}} \ln\left(1 + \frac{Nr^2}{\epsilon}\right)
    \Bigg( \alpha n r L + \frac{12dR^2}{1 - \beta_1} + \frac{2\alpha^2 d L^2 \beta_1}{1 - \beta_1} \Bigg)
  \end{aligned}
\end{equation}

\vspace{0.5cm}
Continuing with Algorithm 1, we can split the proof into two cases:
\begin{itemize}
\item {When \(\beta_3 = \beta_5 = \beta_4 = 0\), the algorithm adopts the form \(\theta_t = \theta_{t-1} - \alpha \frac{m_t}{v_t}\), akin to the AdaGrad algorithm, known for convergence \cite{teddy_2}.}  

\item {For \(\exists j \in \{3,4,5\}: \beta_j > 0\), \(dm_t, tm_5, dv_t,\) and \(tv_t\) depend on \(m_t\) and \(v_t\), converging at some point following the first case. Thus, a \(\delta \in \mathbb{N}\) emerges with \(\epsilon \| \psi_\delta - \psi_{\delta-1} \|_2 \leq \epsilon^*\) for \(\psi \in \{dm, tm, dv, tv\}\). Consequently, beyond this interaction (\(\delta\)), the update rule transforms into \(\forall t > \delta: \theta_t = \theta_{t-1} - \alpha\frac{m_t + c_1}{v_t + c_2}\), with \(c_1, c_2 \in \mathbb{R}^n\). This update rule can be upper-bounded by a constant \(c_3 \in \mathbb{R}^n\), resembling ADAgrad, because the adaptive update roughly aligns with the descent direction. Thus, the convergence at \(\tau \in \mathbb{N}\) implies subsequent steps being smaller, bounded by ADAgrad convergence processes from the same initial configuration.}
\end{itemize}

\noindent Ultimately, by establishing an upper limit to convergence in the second case, we demonstrate convergence in this scenario as well.

\section{Conclusion}
We introduced FAME, a novel optimization algorithm that utilizes TEMA to overcome the fundamental limitations of traditional EMA-based optimizers in accurately tracking gradient trends and mitigating optimization lag. Through extensive empirical evaluation across 30 architectures and a diverse range of computer vision tasks—encompassing classification, detection, and dense prediction—we demonstrated FAME's high versatility, superior robustness, and adaptability. FAME outperformed state-of-the-art optimizers in 86.67\% of evaluated models, consistently maintaining lower variance and stability under diverse conditions. Beyond delivering better performance, FAME demonstrated significant practical advantages by reducing average convergence time by 23.7\%, enhancing training stability, and exhibiting performance less sensitive to specific model architectures or benchmark settings. This versatility underscores FAME’s potential as a reliable optimizer, especially in settings requiring scalability and model independence. With its theoretical basis in hierarchical EMA structures and demonstrated empirical superiority across model-benchmark configurations, FAME enables more efficient and resilient training across a broad spectrum of applications.

{
    \small
    \bibliographystyle{ieeenat_fullname}
    \bibliography{main}
}

% WARNING: do not forget to delete the supplementary pages from your submission 
% \input{supplementary}

\end{document}